\documentclass[12pt]{article}
\usepackage{graphicx} 
\usepackage{amsmath}
\usepackage{biblatex}
\usepackage{authblk}

\title{Brief technical note on linearizing recurrent neural networks (RNNs) before vs after the pointwise nonlinearity}
\author[1,$\dagger$]{Marino Pagan}
\author[2,$\dagger$]{Adrian Valente}
\author[2,*]{Srdjan Ostojic}
\author[3,*]{Carlos D. Brody}

\affil[1]{Simons Initiative for the Developing Brain, University of Edinburgh, Edinburgh, UK}
\affil[2]{Laboratoire de Neurosciences Cognitives et Computationnelles, INSERM U960, Ecole Normale Superieure - PSL Research University, 75005 Paris, France}
\affil[3]{Howard Hughes Medical Institute and Princeton Neuroscience Institute, Princeton University, Princeton NJ, USA}
\affil[$\dagger$]{equal Contribution}
\affil[*]{equal Contribution}

\date{February 2023}

\begin{document}

\maketitle

\begin{abstract}
Linearization of the dynamics of recurrent neural networks (RNNs) is often used to study their properties. The same RNN dynamics can be written in terms of the ``activations" (the net inputs to each unit, before its pointwise nonlinearity) or in terms of the ``activities" (the output of each unit, after its pointwise nonlinearity); the two corresponding linearizations are different from each other. This brief and informal technical note describes the relationship between the two linearizations, between the left and right eigenvectors of their dynamics matrices, and shows that some context-dependent effects are readily apparent under linearization of activity dynamics but not linearization of activation dynamics.
\end{abstract}
\tableofcontents

\def\mvec#1 {\mathbf #1}
\def\mhat#1 {\hat{\mathbf #1} }
\def\gvec#1 {\boldsymbol #1}
\def\gfun {{\rm g}}
\def\bfx {$\bf x$~}
\def\bfr {$\bf r$~}
\def\Fx {{\mathbf F}_x}
\def\Fr {{\mathbf F}_r}

\section{Introduction}

Recurrent neural network (RNN) dynamics can be equivalently expressed in two different forms \cite{Miller2012-ml}.
One form describes the dynamics of the net input, or ``activations" of the units, usually interpreted as  the membrane potential of biological neurons. A second form describes dynamics in terms of the output, i.e. ``activity" or "rate" of the units, often thought of as comparable to spiking rates of biological neurons. A pointwise nonlinearity relates the two, with the activity being the result of the nonlinearity after it is applied to the activation.

Linearization of dynamics is often used to study the properties of dynamical systems. But when considering an RNN, should one linearize the activity dynamics? Or the activation dynamics? The two linearizations produce different linear equations. What is the difference between them and what is the relationship between the two? Do some conclusions depend on which linearization is chosen?

This document explores these questions, and makes the relationship between the two linearizations explicit. The two are related by a simple diagonal linear transform that depends on the gains of each unit. 

We additionally briefly consider effects of the two linearizations when considering context-dependent networks \cite{Mante2013-nt, Pagan2022-ve}, in which each ``context" is defined by a constant vector of inputs to each unit, and point out that a modulation by context of the linearized inputs to the RNN is observable only in the activity space linearization, not in the activation space linearization. 

\section{Two linearizations for the same discrete-time RNN Dynamics}

Consider the standard recurrent neural network equations

\begin{eqnarray}
    \mhat{x} ^{k+1} & = & W\mhat{r} ^k + \mvec{u} ^k  \nonumber \\
    \mhat{r} ^{k+1} & = & \gfun(\mhat{x} ^{k+1}) \label{origdyn}   
\end{eqnarray}
where $\mhat{x} ^k$ represents the vector of unit activations at timepoint $k$, $\gfun()$ is a differentiable and invertible pointwise nonlinear function such as $\tanh()$, $\mhat{r} ^k$ is the vector of unit activities at timepoint $k$, $W$ is a square matrix representing recurrent connection weights, and $\mvec{u} ^k$ is a vector of external inputs at timepoint $k$.

The dynamics of (\ref{origdyn}) can be rewritten entirely in terms of $\bf \hat{x}$. As we do that, let us define the vector-valued dynamics function $\Fx$: 

\begin{equation}
\mvec{F}_x( \mhat{x} ,\mvec{u} ) =  W\gfun(\mhat{x} ) + \mvec{u}  ,
\end{equation}
so that

\begin{eqnarray}
    \mhat{x} ^{k+1} & = & \mvec{F}_x( \mhat{x} ^k,\mvec{u} ^k).  \label{xonlydyn}  
\end{eqnarray}

Similarly, we can define the dynamics function $\Fr
$

\begin{equation}
\mvec{F}_r( \mhat{r} ,\mvec{u} ) \; = \;\gfun(W\mhat{r}  + \mvec{u} )
\end{equation}
and rewrite the dynamics (\ref{origdyn}) entirely in terms of $\bf \hat{r}$,

\begin{eqnarray}
    \mhat{r} ^{k+1} & = & \mvec{F}_r( \mhat{r} ^k,\mvec{u} ^k) \label{ronlydyn}   
\end{eqnarray}

We will consider the effects of linearizing around a fixed point when the dynamics are written in terms of $\Fx$ versus when they are written in terms of $\Fr$.

To begin, consider a point specified by 
\begin{eqnarray}
    \mhat{x} _0 & \nonumber \\
    \mvec{u} _0 & = & \mvec{0} \nonumber \\
    \mhat{r} _0 & = & \gfun(\mhat{x} _0)
\end{eqnarray}
which we choose to be a fixed point of the dynamics (\ref{origdyn}), i.e., it is such that 
\begin{eqnarray}
    \mhat{x} _0 & = &\mvec{F}_x( \mhat{x} _0,\mvec{u}_0). \label{xfp} 
\end{eqnarray}

Linearizing $\Fx$ around that fixed point, we obtain

\begin{eqnarray}
    \mhat{x} ^{k+1} = \Fx(\mhat{x} ^k, \mvec{u} ^k) \approx \; \Fx(\mhat{x} _0) \; + \; \frac{\partial \Fx}{\partial \mhat{x} } (\mhat{x} ^k - \mhat{x} _0)  \; + \; \frac{\partial \Fx}{\partial \mvec{u} } \mvec{u} ^k
\end{eqnarray}

Inserting Eq.~\ref{xfp}, we obtain

\begin{eqnarray}
    \mhat{x} ^{k+1} - \mhat{x} _0 = \frac{\partial \Fx}{\partial \mhat{x} } (\mhat{x} ^k - \mhat{x} _0) \; + \; \frac{\partial \Fx}{\partial \mvec{u} } \mvec{u} ^k.
\end{eqnarray}
Changing variables to 
\begin{eqnarray}
    \mvec{x} ^k = \mhat{x} ^k - \mhat{x} _0, \label{xvarchange}
\end{eqnarray}
we arrive at

\begin{eqnarray}
    \mvec{x} ^{k+1} = \frac{\partial \Fx}{\partial \mhat{x} } \mvec{x} ^k \; + \; \frac{\partial \Fx}{\partial \mvec{u} } \mvec{u} ^k.  \label{linX}
\end{eqnarray}
In index notation, the two matrices involved in (\ref{linX}) are

\begin{eqnarray}
    \left[\frac{\partial \Fx}{\partial \mhat{x} } \right]_{ij}   & = & W_{ij} \gfun'(\hat{x}_{0j}) \label{indexX}\\
\vspace*{-0.5 cm} \nonumber \\
    \left[ \frac{\partial \Fx}{\partial \mvec{u} } \right]_{ij}  & = & \delta_{ij} \label{indexU}
\end{eqnarray}

Let us define a diagonal matrix $D$, i.e., with zeros on all the non-diagonals, and entries along the diagonal that are each a function of the $j^{\rm th}$ element of $\mhat{x} _0$ : 

\begin{eqnarray}
D_{jj} = \gfun'(\hat{x}_{0j}).    
\end{eqnarray}

Since its elements are the gains of $\gfun$ for each element of $\hat{x} _0$, we will call this matrix \textbf{the diagonal gain matrix} $D$. Then, in matrix notation, we can use $D$ to rewrite the linearized dynamics (\ref{linX}) as 

\begin{center}
\noindent\fbox{%
    \parbox{0.8\textwidth}{%
        \begin{eqnarray}
            \mvec{x} ^{k+1} & = & W D \, \mvec{x} ^k + \mvec{u} ^k 
            \label{xinstance}
        \end{eqnarray}
    }%
}
\end{center}

The second linearization is obtained by differentiating equation (\ref{ronlydyn}) with respect to $\mhat{r} $ and changing variables to 

\begin{eqnarray}
    \mvec{r} ^k = \mhat{r} ^k - \mhat{r} _0.  \label{rvarchange}
\end{eqnarray}

This requires the derivatives

\begin{eqnarray}
    \left[\frac{\partial \Fr}{\partial \mhat{r} } \right]_{ij}  & = & \gfun'(\hat{x}_{0i}) W_{ij}  \label{indexR}\\
\vspace*{-0.5 cm} \nonumber \\
    \left[\frac{\partial \Fr}{\partial \mvec{u} } \right]_{ij}  & = & \gfun'(\hat{x}_{0i}) \delta_{ij} \label{indexU_r}
\end{eqnarray}

which we rewrite in matrix notation as 

\begin{center}
\noindent\fbox{%
    \parbox{0.8\textwidth}{%
        \begin{eqnarray}
            \mvec{r} ^{k+1} & = & D W \mvec{r} ^k + D \mvec{u} ^k 
            \label{rinstance}
        \end{eqnarray}
    }%
}
\end{center}

The two linear dynamical systems (\ref{xinstance}) and (\ref{rinstance}) might appear at first sight to be quite disparate.  $DW$ represents a scaling of the \textit{rows} of $W$ by the diagonal elements of $D$, while $WD$ represents a scaling of the \textit{columns} of $W$ by the diagonal elements of $D$. The results of the two scalings could be quite different, suggesting that different conclusions might be drawn from analyzing $WD$ versus analyzing $DW$, even though they are both linearizations of the same dynamics around the same fixed point.

But this is not the case. The two equations describe dynamics in terms of different variables, $\mvec{x} $ and $\mvec{r} $ , but are in fact intimately related. If we express the dynamics in terms of the same variable,  the two different linearizations lead to identical trajectories.

To relate the variables $\mvec{x} $ and $\mvec{r}$, consider a linearization of $\gfun()$ around $\hat{\mathbf x}_0$ so that 

\begin{eqnarray}
\mhat{r} & \approx & \gfun(\mhat{x} _0) + \gfun'(\mhat{x} _0)(\mhat{x} - \mhat{x} _0)
\end{eqnarray}

Then, given that $\mhat{r} _0 = \gfun(\mhat{x} _0)$, and using the variable changes (\ref{xvarchange}) and (\ref{rvarchange}), we can find the map relating $\mvec{r}~$ and $\mvec{x}$:

\begin{eqnarray}
\mvec{r} & \approx & \gfun'(\mhat{x} _0) \, \mvec{x} \; = \; D \mvec{x} \label{rxmap}
\end{eqnarray}

This makes it plain that the two equations (\ref{xinstance}) and (\ref{rinstance}) are equivalent, related through the map in (\ref{rxmap}). That is, we can take equation (\ref{xinstance}), multiply it on the left by the gain matrix $D$, and obtain equation (\ref{rinstance}):

\begin{eqnarray}
    \mvec{x} ^{k+1} & = & W D \, \mvec{x} ^k + \mvec{u} ^k \nonumber \\
    D \mvec{x} ^{k+1} & = & D W D \, \mvec{x} ^k + D \mvec{u} ^k \nonumber \\
    \mvec{r} ^{k+1} & = & D W \, \mvec{r} ^k + D \mvec{u} ^k 
\end{eqnarray}

This means that if we take a trajectory of  points $\mvec{x} ^k$ produced by the linearization of $\Fx$ in (\ref{xinstance}), and map each $\mvec{x} ^k$  onto its corresponding $\mvec{r} ^k$ using (\ref{rxmap}), we will get exactly the set of $\mvec{r} ^k$ that the linearization of $\Fr$ in (\ref{rinstance}) would have produced.  The two linearizations describe the same trajectories and thus the same dynamics, albeit mapped onto each other through $D$, as in (\ref{rxmap}).

%
%
%

\section{Left and right eigenvectors of the dynamics matrices}

As we have described, (\ref{xinstance}) and (\ref{rinstance}) are two views of the same dynamical trajectories. But they have different linearized dynamics matrices, respectively $WD$ and $DW$, which in general have different eigendecompositions. The right and left eigenvectors of linearized dynamics matrices determine many features of the dynamics, but as shown above, the dynamics are independent of the chosen linearization. This suggests that the eigendecompositions of the two matrices should be closely related, and here we show that indeed they  are.

Let $W$ be a square matrix and $D$ be a diagonal matrix of the same size as $W$.

Let $\mvec{s} _r^T$ be a \textbf{left} eigenvector of matrix $DW$, with corresponding eigenvalue $\lambda$. In other words,

\begin{eqnarray}
    \mvec{s} _r^T DW = \lambda \mvec{s} _r^T
\end{eqnarray}

Multiplying on the right by $D$ we obtain

\begin{eqnarray}
    \mvec{s} _r^T DWD = \lambda \mvec{s} _r^T D
\end{eqnarray}

which means that the vector $\mvec{s} _r^T D$ is a left eigenvector of the matrix $WD$, with eigenvalue $\lambda$.  

In other words,

\begin{center}
\noindent\fbox{%
    \parbox{0.9\textwidth}{%
    If $\mvec{s} _r^T$ is a \textbf{left} eigenvector of $DW$ with eigenvalue $\lambda$, then
        \begin{eqnarray}
        \mvec{s} _x^T = \mvec{s} _r^T D
        \end{eqnarray}
    is a corresponding \textbf{left} eigenvector of $WD$, also with eigenvalue $\lambda$.
    }%
}
\end{center}

Similarly, let $\gvec{\rho} _x$ be a \textbf{right} eigenvector of $WD$, with eigenvalue $\lambda$.  That is,

\begin{eqnarray}
    WD \gvec{\rho} _x = \lambda \gvec{\rho} _x
\end{eqnarray}

Multiplying on the left by $D$ we obtain

\begin{eqnarray}
    D W D \gvec{\rho} _x = \lambda D \gvec{\rho} _x
\end{eqnarray}

which means that the vector $D \gvec{\rho} _x$ is a right eigenvector of the matrix $DW$, with eigenvalue $\lambda$.  

In other words,

\begin{center}
\noindent\fbox{%
    \parbox{0.9\textwidth}{%
    If $\gvec{\rho} _r$ is a \textbf{right} eigenvector of $DW$ with eigenvalue $\lambda$, then
        \begin{eqnarray}
        \gvec{\rho} _x = D^{-1} \gvec{\rho} _r
        \end{eqnarray}
    is a corresponding \textbf{right} eigenvector of $WD$, also with eigenvalue $\lambda$.
    }%
}
\end{center}

These relationships imply that the dot product between left and right eigenvectors is preserved:

\begin{center}
\noindent\fbox{%
    \parbox{0.5\textwidth}{%
\begin{eqnarray}
    \mvec{s} _x^T \cdot \gvec{\rho} _x & = & \nonumber \\
    & = & \mvec{s} _r^T D \cdot D^{-1} \gvec{\rho} _r \nonumber \\
    & = & \mvec{s} _r^T \cdot \gvec{\rho} _r \nonumber
\end{eqnarray}
    }%
}
\end{center}

Note that, except for the case when W is rank 1, the relationship between the eigenvectors of $W$ and the eigenvectors of $WD$ or $DW$ is in general non-trivial.

%
%

\section{Linearizations and context-dependence of input vectors}

Any given RNN will be defined by its weight matrix $W$, and trajectories on it will be induced by inputs $\mvec{u} ^k$, where $k$ indexes timepoints. We define $\mvec{u} ^k=0$ for $k<0$, and consider the case where the network is simulated over multiple different ``runs" or ``trials", each of which begins at a timepoint $k<<0$, and evolves to some timepoint $k>0$.  Let us now consider a situation in which there are additional inputs to the units of the network, constant in time during each run, but potentially different across different runs.  That is, during each run $R$, the dynamical equations are

\begin{eqnarray}
    \mhat{x} ^{k+1} & = & W\mhat{r} ^k + \mvec{u} ^k  + \mvec{c} _R \nonumber \\
    \mhat{r} ^{k+1} & = & \gfun(\mhat{x} ^{k+1}) \label{contextdyn}   
\end{eqnarray}

The inputs $\mvec{c} _R$ define what we will call \textit{context R}. 

Let us further suppose that before timepoint $k=0$ of each run in context $R$, and before any inputs $\mvec{u}~$ are non-zero in that run, the network has settled into a fixed-point determined by $\mvec{c} _R$. This fixed-point will be such that

\begin{eqnarray}
    \mhat{x} _0^R & = & W \gfun(\mhat{x} _0(\mvec{c} _R)) + \mvec{c} _R \label{cxfp}   
\end{eqnarray}

and will have a corresponding gain matrix $D_R$ whose diagonal entries are the elements of $\gfun'(\mhat{x} _0(\mvec{c} _R)).$

Following (\ref{xinstance}) and (\ref{rinstance}), let us define the linearization of the network for context $R$ to be the linear network with dynamics

\begin{eqnarray}
    \mvec{x} ^{k+1} & = & W D_R \, \mvec{x} ^k + \mvec{u} ^k \label{activationSpaceInstance} \\
        & {\rm and} & \nonumber \\
    \mvec{r} ^{k+1} & = & D_R W \, \mvec{r} ^k + D_R \mvec{u} ^k \label{instanceDynamics}
\end{eqnarray}

Differences between two contexts $A$ and $B$ in how a network behaves will then correspond to different instantiations of the network, one determined by the gain matrix $D_A$, the other by the gain matrix $D_B$.

\begin{center} 
\noindent\fbox{%
    \parbox{0.9\textwidth}{%
    Notice that context-dependent modulation of the linearized input $\mvec{u}~$ is observable only in the activity space linearization (\ref{instanceDynamics}) (where the linearized input is $D_R \mvec{u},~$ and thus depends on the gain matrix $D_R$). In the activation space  linearization (\ref{activationSpaceInstance}), the linearized input is always $\mvec{u},~$ independent of $D_R$. 
    }%
}
\end{center}

Context-dependent input modulation of recurrent networks with a fixed input vector $\mvec{u}~$ is studied, for example, in \cite{Maheswaranathan2020-fl}, who utilize activity space linearization (\ref{instanceDynamics}) for this purpose: the linearized inputs $D_R \mvec{u}~$ depend on context through $D_R$. In contrast, \cite{Mante2013-nt} used activation space linearization (\ref{activationSpaceInstance}) when studying context dependence of RNN dynamics with fixed input vectors, and therefore did not study context-dependent input modulation.

\section{Conclusion}

In a recurrent neural network, the linear dynamics that result from linearization in activation space, and those that result from linearization in activity space, are different. Nevertheless, the two linear dynamics describe the same underlying trajectories, albeit mapped onto each other through a scaling given by the gain of each of the network's units.  

Despite this close relationship between the two linearizations, the two are not interchangeable. In particular, context-dependent modulations of external inputs that follow from context-dependent changes in unit gains are directly observable as input modulations in the activity space linearization, but not in the activation space linearization.

\printbibliography

\end{document}